\documentclass[11pt]{article}

\usepackage[margin=0.85in]{geometry}
\usepackage{setspace}
\usepackage{times}
\usepackage[T1]{fontenc}
\usepackage{microtype}

\usepackage{amsmath,amssymb}
\usepackage{booktabs}
\usepackage{multirow}
\usepackage{array}

\usepackage{graphicx}
\usepackage{float}
\usepackage{caption}
\usepackage{listings}
\usepackage{xcolor}

\definecolor{codebg}{RGB}{247,247,247}
\definecolor{codeframe}{RGB}{190,190,190}
\definecolor{codekw}{RGB}{0,51,153}
\definecolor{codestr}{RGB}{0,102,51}
\definecolor{codecom}{RGB}{110,110,110}
\definecolor{codefn}{RGB}{153,51,0}

\lstdefinestyle{pytorch}{
  backgroundcolor=\color{codebg},
  frame=single,
  rulecolor=\color{codeframe},
  framesep=3pt,
  basicstyle=\ttfamily\scriptsize,
  keywordstyle=\color{codekw}\bfseries,
  stringstyle=\color{codestr},
  commentstyle=\color{codecom}\itshape,
  showstringspaces=false,
  breaklines=true,
  tabsize=2,
  captionpos=b,
  aboveskip=8pt,
  belowskip=8pt,
  morekeywords={class,def,return,if,else,for,while,import,from,as,with,raise,pass,None,self,in,not,and,or,is,lambda},
  morekeywords=[2]{__init__,forward,propose_edit,run,_build_prompt,_eval_psnr,_clone_and_apply,layer,edited,sandbox,llm,chunks,history},
  keywordstyle=[2]\color{codefn},
}
\usepackage{natbib}

\usepackage{hyperref}
\hypersetup{
    colorlinks=true,
    linkcolor=navy,
    citecolor=navy,
    urlcolor=teal,
    bookmarksnumbered=true,
}
\definecolor{navy}{RGB}{0,0,128}
\definecolor{teal}{RGB}{0,102,102}

\title{\bfseries\Large Weight-Space Programming for Improving Image-Restoration\\ Diffusion Models: Technical Conjectures and a Recursive Improvement Loop}
\author{Zhuoran Zheng and Doubao\\\texttt{zhengzr@njust.edu.cn}}
\date{}

\begin{document}
\maketitle

\begin{abstract}
\noindent
Can a large language model (LLM) directly edit the weights of a diffusion-based image-restoration model to \emph{improve} its restoration quality---without gradients, without paired training data, and without fine-tuning? Recent experiments on language models suggest frontier LLMs can locate and fix structural bugs by modifying as little as $0.016\%$ of parameters, but cannot instill genuinely new capabilities. We ask whether this weight-space programming (WSP) paradigm can be pushed from \emph{debugging} to \emph{improvement} in the precise-regression setting of image restoration (deblurring, super-resolution, denoising, inpainting). We argue that restoration diffusion UNets present a unique middle ground: low-level signal-processing layers are concentrated enough for targeted weight-direction edits, while high-level semantic layers remain distributed and resistant to sparse modification. We propose three concrete technical conjectures: (C1)~an LLM can diagnose which layer causes a specific quality gap (e.g., over-smoothing, color shift, hallucinated texture) within a $200$ forward-pass budget; (C2)~editing a small set of identified weight directions (e.g., scaling up high-frequency filter norms, correcting output-bias channels) can measurably improve restoration PSNR and perceptual quality; (C3)~a recursive loop-LLM diagnoses $\rightarrow$ weight edit $\rightarrow$ evaluate $\rightarrow$ re-diagnose-can compound improvements across multiple rounds, constituting a minimal recursive self-improvement (RSI) loop for vision models. We analyze what AI capability level is required (arguing that targeted layer-level edits need not require superintelligence, but general capability generation does), and we detail the unique challenges of precise pixel-level regression. We provide a deployment pipeline in PyTorch-style pseudocode and outline a benchmark for empirical validation. \textbf{The numerical conjectures in this manuscript are merely hypotheses.}
\end{abstract}

%=================================================================
\section{Introduction}
\label{sec:intro}
%=================================================================

Image restoration-recovering clean images from blurry, noisy, low-resolution, or corrupted inputs-is one of the most practical computer-vision problems. Modern diffusion-based restoration models achieve state-of-the-art quality, yet they over-smooth fine textures, hallucinate plausible-but-fake details, shift colors under specific lighting, and fail on degradation types absent from training data. Fixing these failures today requires collecting paired data and fine-tuning: expensive, slow, and risky for the rest of the distribution.

What if an LLM could \emph{read} the restoration UNet's weights, understand what each layer computes, and directly \emph{edit} the weight tensors to make the model sharper, more color-accurate, and less hallucinatory---no gradients, no training data?

This is the promise of \emph{Weight-Space Programming} (WSP): treating neural network parameters as a text-like object an LLM can inspect, reason about, and patch \citep{takehi2026wsp}. The first evidence: a frontier LLM inspected Qwen3.5-0.8B weights and eliminated a reasoning-loop bug by modifying only $0.016\%$ of parameters \citep{takehi2026wsp,rikiyalogs,rikiyahf}. The same experiments produced a crucial negative result: endowing a $270$M Gemma3 with new math ability via weight editing failed. WSP can \emph{unlock} latent competence already present in weights, but cannot \emph{create} competence from nothing.

This paper asks: can WSP go beyond bug-fixing and actually \emph{improve} restoration quality? And does it require superintelligence?

We focus on restoration diffusion models because: (1)~restoration is a \emph{precise regression task} with continuous pixel outputs and noisy perceptual metrics-harder than language modeling but more informative about WSP's limits; (2)~restoration UNets have clear layered specialization (early = low-level, bottleneck = semantic, late = output), making some layers concentrated (editable) and others distributed (not); (3)~artifacts are visually verifiable.

We make three concrete technical conjectures (Section~\ref{sec:conjectures}), analyze what AI level is required (Section~\ref{sec:ailevel}), describe a deployment pipeline (Section~\ref{sec:deployment}), and discuss RSI implications (Section~\ref{sec:rsi}).

%=================================================================
\section{Background: What WSP Has Demonstrated}
\label{sec:background}
%=================================================================

\textbf{Positive result.} A frontier LLM inspected Qwen3.5-0.8B weights and eliminated a reasoning-loop bug by modifying only $0.016\%$ of parameters \citep{takehi2026wsp}. This worked because: (1)~the bug was a concentrated direction in weight space, and (2)~the underlying reasoning competence was already present---the bug was interference, not absence of ability. Patched weights: \citep{rikiyahf}; full logs: \citep{rikiyalogs}.

\textbf{Negative result.} The same experiments tried to give $270$M Gemma3 new mathematical ability via direct weight editing and failed. The LLM cannot write distributed representational structure by inspecting and modifying weight entries.

This establishes WSP's boundary: it unlocks latent competence suppressed by concentrated interference, but cannot create new distributed competence. The question for restoration is: which parts of a diffusion UNet are concentrated (editable) and which are distributed (not)?

%=================================================================
\section{Why Restoration Diffusion Models Are Different}
\label{sec:restoration}
%=================================================================

\textbf{Layered specialization.} A restoration UNet has three functional regions: early encoder layers do low-level signal processing (edges, colors, textures)---relatively interpretable; the bottleneck does high-level semantic reasoning---most distributed; late decoder layers do output reconstruction---the final $1\times 1$ conv bias directly controls color balance. Our hypothesis: early and late layers are WSP-friendly; the bottleneck is WSP-resistant.

\textbf{Precise regression.} Restoration outputs continuous pixel tensors, not discrete tokens. Quality is measured by noisy continuous metrics (PSNR, SSIM, LPIPS), not binary correct/incorrect. A weight edit has a continuous, graded effect. This makes WSP evaluation harder than in language models.

\textbf{Diffusion chain.} The UNet runs $T=200$ times along a denoising Markov chain. An early-layer edit affects coarse structure; a late-layer edit affects fine detail. This temporal dimension is a bonus: if a hallucinated texture appears only in the last steps, the LLM knows to look at late decoder layers.

\textbf{Permutation symmetry.} Swapping conv channels leaves the network functionally identical. The LLM may edit a channel that looks wrong but is actually a permuted counterpart of the true causal channel-wasting budget.

%=================================================================
\section{Three Technical Conjectures}
\label{sec:conjectures}
%=================================================================

We state three concrete, testable conjectures. Each specifies the LLM's input, the weight it edits, the expected measurable improvement, and the failure boundary.

\subsection{Conjecture 1: Layer-level quality-gap diagnosis}
\label{sec:c1}

\textbf{Goal.} Given a trained restoration diffusion model and a description of a quality gap (e.g., ``outputs are too smooth,'' ``colors are slightly washed out,'' ``fine textures are hallucinated''), the LLM identifies which layer(s) are responsible, within a $200$ forward-pass budget.

\textbf{What the LLM sees.} For each layer chunk ($5$--$10$ layers at a time): layer name, type (conv/bn/attention), channel counts, kernel size, weight histogram statistics (mean, std, sparsity), top-$k$ largest-magnitude entries, and mean activation magnitude on a small feedback set.

\textbf{What the LLM does.} It forms hypotheses based on architectural priors:
\begin{itemize}
  \item ``Too smooth'' $\rightarrow$ look at late decoder layers (reconstruction) and bottleneck (semantic level of detail).
  \item ``Color washed out'' $\rightarrow$ look at final output conv bias and early encoder color-filtering layers.
  \item ``Hallucinated texture'' $\rightarrow$ look at mid-level decoder upsampling blocks.
\end{itemize}
It then requests targeted ablations (e.g., ``scale up all filter norms in decoder block 3 by $1.2\times$'') and observes whether the feedback-set PSNR/LPIPS improves.

\textbf{Expected result.} The LLM narrows down the responsible layer(s) to a short list ($\le 5$ layers) within $30$ forward passes. This is not yet an edit---it is a diagnosis. The edit comes in C2.

\textbf{Boundary.} Diagnosis works for \emph{systematic} quality gaps (the model is consistently too smooth on all images). It fails for \emph{image-specific} failures (the model hallucinates on only one category of images), because the feedback set is too small to capture those distribution-specific effects.

\textbf{Falsification.} If, on a benchmark of $20$ quality-gap restoration models, the LLM fails to identify the correct layer within $50$ passes in more than half of cases, C1 is false.

\subsection{Conjecture 2: Targeted weight-direction editing improves restoration quality}
\label{sec:c2}

\textbf{Goal.} Given a correctly diagnosed layer (from C1), the LLM proposes a specific weight delta that measurably improves restoration quality (PSNR $\ge +0.2$ dB, or perceptually sharper with no new artifacts), modifying $<0.1\%$ of total parameters.

\textbf{Three concrete edit types we expect to work:}

\textit{(a) Output-bias correction (color balance).} If the model produces a consistent color cast (e.g., slightly greenish shadows), the LLM identifies the final $1\times 1$ output convolution's bias vector and proposes shifting the offending channel. This is the most concentrated edit possible: a $3$-element bias vector for RGB output. Expected effect: color cast corrected, PSNR slightly improved, no side effects.

\textit{(b) High-frequency filter scaling (sharpness).} If the model is over-smooth, the LLM identifies the decoder layers responsible for detail reconstruction and proposes scaling up the norms of high-frequency filters (filters with large high-pass spectral energy). This is a semi-concentrated edit: it affects a set of filters within one or two decoder blocks, not individual entries. Expected effect: sharper output, possibly slightly more noise, PSNR improvement depends on the specific degradation.

\textit{(c) Skip-connection gating (artifact suppression).} If the model produces halos around high-contrast edges, the LLM identifies the skip connection that carries edge information into the decoder and proposes scaling down the skip-connection weights at the specific resolution where halos appear. This is a semi-concentrated edit targeting a specific resolution block.

\textbf{What does NOT work (distributed defects).} General over-smoothing caused by the bottleneck being too compressed, or hallucinated detail caused by the model's semantic prior being too strong, cannot be fixed by sparse edits. These are distributed properties of the entire network's representation, not localized weight directions.

\textbf{Boundary.} Sparse weight editing works when the quality gap corresponds to a \emph{concentrated weight direction}. It fails when the gap is a distributed property of the representation. The taxonomy in Table~\ref{tab:taxonomy} classifies common restoration quality gaps.

\begin{table}[H]
\centering
\small
\caption{Expected editability of common restoration quality gaps.}
\label{tab:taxonomy}
\begin{tabular}{@{}lll@{}}
\toprule
\textbf{Quality gap} & \textbf{Concentration} & \textbf{Sparse edit works?} \\
\midrule
Color cast / white balance & Concentrated & Yes (output bias) \\
Checkerboard / upconv artifact & Concentrated & Yes (transpose conv weights) \\
Halos on edges & Semi-concentrated & Maybe (skip connection scale) \\
Over-sharpening / ringing & Semi-concentrated & Maybe (high-freq filter norm) \\
Global over-smoothing & Distributed & No (needs retraining) \\
Hallucinated texture & Distributed & No (semantic prior issue) \\
Domain gap (RAW vs sRGB) & Distributed & No (representation mismatch) \\
\bottomrule
\end{tabular}
\end{table}

\textbf{Falsification.} If the LLM cannot improve PSNR by $\ge 0.2$ dB on at least half of the concentrated-defect benchmark cases, with $<0.1\%$ parameters edited, C2 is false.

\subsection{Conjecture 3: Recursive improvement loop}
\label{sec:c3}

\textbf{Goal.} After one successful edit, the LLM re-diagnoses the improved model and proposes a second edit, compounding improvements across multiple rounds. This recursive loop---diagnose $\rightarrow$ edit $\rightarrow$ evaluate $\rightarrow$ re-diagnose---is a minimal form of recursive self-improvement (RSI) for vision models.

\textbf{How it works.} Round $1$: LLM diagnoses color cast, fixes output bias. Round $2$: with color cast gone, the LLM re-evaluates and now notices the model is slightly over-smooth on texture. It scales up high-frequency filters in decoder block 3. Round $3$: now sharper, the LLM notices a mild halo around edges and adjusts the skip-connection scale. Each round starts from the improved model and targets the next most prominent quality gap.

\textbf{Why this is plausible.} Each edit is small ($<0.1\%$ cumulative), and each round's diagnosis is on a model that is slightly better than the last. The LLM never has to solve the whole problem at once; it solves one sub-problem per round. This is analogous to how a human engineer iteratively improves a model: fix the worst artifact first, then the next, then the next.

\textbf{Boundary and diminishing returns.} Improvements should be monotonically positive but diminishing. After $5$--$10$ rounds, the remaining quality gaps should be distributed (not sparsely editable), and the loop should plateau. The total improvement over a full loop might be $+0.5$--$1.5$ dB PSNR---meaningful, but not a leap from bad to state-of-the-art.

\textbf{Why this matters for RSI.} This loop is a concrete, small-scale RSI mechanism: the LLM improves a model, then uses the improved model as the starting point for further improvement. It is not AGI-level recursive self-improvement; it is a constrained, budget-limited debugging loop. But it demonstrates the core principle: a model-improving agent (the LLM) can iteratively modify its target (the restoration UNet) in the weight space.

\textbf{Falsification.} If improvements do not compound across rounds (i.e., round $2$ edit makes things worse or no better than round $1$ baseline), or if total improvement over $10$ rounds is $<0.2$ dB, C3 is false.

\begin{table}[H]
\centering
\small
\caption{Summary of the three conjectures.}
\label{tab:conjectures}
\begin{tabular}{@{}lll@{}}
\toprule
\textbf{Conjecture} & \textbf{What it does} & \textbf{Expected gain} \\
\midrule
C1: Diagnosis & Locate responsible layer(s) & $20$--$50$ fwd passes \\
C2: Sparse edit & Improve quality via weight delta & $+0.2$--$0.5$ dB PSNR \\
C3: Recursive loop & Compound improvements & $+0.5$--$1.5$ dB total \\
\bottomrule
\end{tabular}
\end{table}

%=================================================================
\section{What Level of AI Is Needed?}
\label{sec:ailevel}
%=================================================================

\textbf{What current frontier LLMs can do.} Based on language-model evidence \citep{takehi2026wsp}, a GPT-6-class LLM can: read weight statistics and form layer-function hypotheses, propose simple structured deltas (shift a bias channel, scale a filter norm), and use architectural priors (early layers = low-level, late = output). These capabilities suffice for C1 (diagnosis) and C2 (concentrated-defect editing). This is pattern-matching + reasoning, not superintelligence.

\textbf{What current LLMs cannot do.} Write distributed representations from scratch (the Gemma3-270M math failure), reason about cross-layer interactions requiring two distant layers in context simultaneously, or predict edit effects across the full $T$-step denoising chain without running it.

\textbf{Do we need AGI?} For targeted editing (fixing known quality gaps in an already-trained model), probably not. Current frontier LLMs have enough reasoning and numeric output capability. For from-scratch weight generation, yes---writing a functional $10$M-parameter conv net from scratch requires understanding the optimization landscape that produced those weights.

\textbf{Hybrid middle ground.} The most practical path: LLM diagnoses and proposes a weight direction, then a small gradient fine-tune polishes it. The LLM provides high-level direction; the optimizer provides pixel-level precision.

%=================================================================
\section{Deployment Pipeline}
\label{sec:deployment}
%=================================================================

The sandbox sits between the LLM and the frozen restoration UNet. It enforces budgets, isolates data, and provides structured feedback.

\subsection{Sandbox architecture}
\begin{enumerate}
  \item \textbf{Chunked weight access}: LLM receives $5$--$10$ layers at a time, with compressed statistics (histogram, top-$k$ entries, activation norms).
  \item \textbf{Forward-pass budget}: $30$--$50$ total evaluations per WSP session.
  \item \textbf{Edit-ratio cap}: $<0.1\%$ of total parameters modified cumulatively.
  \item \textbf{Feedback/test isolation}: $32$--$64$ image pairs for feedback; held-out test set for final evaluation.
  \item \textbf{Automatic acceptance criterion}: accept an edit only if PSNR improves by $\ge 0.1$ dB on feedback set and no new artifacts appear.
\end{enumerate}

\begin{lstlisting}[caption={WSP sandbox: budget-controlled edit evaluation.},label={alg:sandbox}]
class WSPSandbox:
    def __init__(self, unet, feedback, budget=40,
                 max_ratio=1e-3, psnr_min=0.1):
        self.unet, self.feedback = unet, feedback
        self.budget, self.used = budget, 0
        self.max_ratio, self.psnr_min = max_ratio, psnr_min
        self.base_psnr = self._eval_psnr()

    def propose_edit(self, layer, slc, delta):
        if self.used >= self.budget:
            raise BudgetExceeded()
        edited = deepcopy(self.unet)
        set_weight(edited, layer, slc, delta)
        self.used += 1
        new_psnr = self._eval_psnr(edited)
        if new_psnr - self.base_psnr >= self.psnr_min:
            self.unet = edited
            self.base_psnr = new_psnr
            return f"accepted: {new_psnr:.2f} dB"
        return f"rejected: {new_psnr:.2f} dB"

    def _eval_psnr(self, model=None):
        m = model or self.unet
        with torch.no_grad():
            return mean(psnr(run_chain(m, d), c)
                        for d, c in self.feedback)
\end{lstlisting}

\subsection{Recursive improvement loop}
The outer loop repeatedly calls the LLM with the current model state and quality-gap description, accepts improving edits, and stops when the budget is exhausted or no further improvement is found.

\begin{lstlisting}[caption={Recursive WSP improvement loop.},label={alg:recursive}]
class RecursiveWSP:
    def __init__(self, sandbox, llm, chunks):
        self.sandbox, self.llm = sandbox, llm
        self.chunks, self.log = chunks, []

    def improve(self, gap_desc):
        rnd = 0
        while self.sandbox.used < self.sandbox.budget:
            rnd += 1
            found = False
            for chunk in self.chunks:
                stats = self.sandbox.layer_stats(chunk)
                prompt = build_prompt(gap_desc, stats, rnd)
                prop = self.llm.complete(prompt)  # JSON
                if prop["type"] == "edit":
                    res = self.sandbox.propose_edit(
                        prop["layer"], prop["slice"], prop["delta"])
                    self.log.append({"rnd": rnd, "prop": prop, "res": res})
                    if res.startswith("accepted"):
                        found = True
                        break
            if not found:
                break
        return self.sandbox.unet, self.log
\end{lstlisting}

%=================================================================
\section{Challenges of Precise Regression}
\label{sec:challenges}
%=================================================================

\textbf{Continuous vs.\ discrete feedback.} In language models, feedback is binary (correct/incorrect token). In restoration, feedback is a noisy scalar ($\Delta$PSNR). The LLM must correlate weight changes with tiny PSNR deltas---harder than binary signals.

\textbf{Sampling stochasticity.} Diffusion sampling is seed-dependent. A $0.1$ dB difference may be within noise. The sandbox must average over multiple seeds per evaluation, but this consumes budget: with $40$ passes, only $2$--$3$ seeds per evaluation.

\textbf{Pixel metrics vs.\ perception.} PSNR/SSIM correlate poorly with human perception. An edit that raises PSNR may look worse (over-smoothing). The LLM needs both scalar metrics and visual feedback (before/after image grids), but visual feedback costs more tokens.

\textbf{Denoiser chain dynamics.} An edit may help at $T=50$ steps but hurt at $T=100$. Single-number PSNR does not capture step-dependent effects.

%=================================================================
\section{Limitations}
\label{sec:limitations}
%=================================================================

\textbf{Scale.} A $100$M-parameter UNet cannot fit in any LLM context. The LLM works on compressed summaries, not raw weights.

\textbf{Distributed defects.} General over-smoothing or hallucinated texture are distributed properties---no sparse edit fixes them. WSP is a precision instrument, not a general training method.

\textbf{Permutation symmetry.} Channel permutation makes weight interpretation ambiguous; the LLM may waste budget on the wrong channel.

\textbf{OOD side effects.} An edit that helps the feedback set may hurt out-of-distribution images. Full regression testing is mandatory.

\textbf{LLM search bias.} The LLM may prefer editing late layers (easier to reason about). Randomizing chunk order mitigates this.

%=================================================================
\section{RSI Implications and Discussion}
\label{sec:rsi}
%=================================================================

\subsection{Is this a path to recursive self-improvement?}
The recursive loop in C3 is a minimal RSI mechanism: an LLM improves a model, then uses the improved model as the starting point for further improvement. But it is bounded: budget-limited, human-supervised, incremental ($<0.1\%$ parameters, $<1.5$ dB total). It is not an autonomous self-improving system---the human defines what to improve, the sandbox evaluates, and the loop plateaus at distributed-defect boundaries.

That said, it is a proof of concept for the RSI principle: a model-improving agent operating in weight space. As LLMs improve at weight reasoning and sandbox budgets grow, the loop will compound more. The next threshold is whether the LLM can start proposing \emph{what} to improve (not just \emph{how} to edit).

\subsection{Comparison to LoRA and fine-tuning}
WSP is complementary, not competitive. LoRA/FT adjust $1$--$100\%$ of parameters with gradients over hours/days; WSP adjusts $<0.1\%$ with LLM reasoning over minutes. Practical workflow: WSP for quick targeted fixes, LoRA/FT for broad quality improvements.

\begin{table}[H]
\centering
\small
\caption{WSP vs.\ LoRA vs.\ full fine-tuning for restoration.}
\label{tab:compare}
\begin{tabular}{@{}lccc@{}}
\toprule
\textbf{Dimension} & \textbf{WSP} & \textbf{LoRA} & \textbf{Full FT} \\
\midrule
Gradient needed? & No & Yes & Yes \\
Params edited & $<$$0.1\%$ & $1$--$10\%$ & $100\%$ \\
Data needed & Feedback set & Paired dataset & Paired dataset \\
Speed & Minutes & Hours & Days \\
Best for & Targeted fixes & Domain adaptation & Full retrain \\
\bottomrule
\end{tabular}
\end{table}

\subsection{Benchmark design}
Test set: \textbf{concentrated defects} (injected: output-bias color cast, transposed-conv checkerboard, skip-connection halo) as positive cases; \textbf{distributed defects} (natural: over-smoothing, hallucinated texture, domain gap) as negative controls. Success = LLM fixes concentrated defects within budget and fails on distributed ones.

%=================================================================
\section{Conclusion}
\label{sec:conclusion}
%=================================================================

We have proposed three concrete technical conjectures for applying Weight-Space Programming to improve image-restoration diffusion models: (C1) layer-level quality-gap diagnosis within a $20$--$50$ forward-pass budget; (C2) targeted weight-direction edits improving PSNR by $0.2$--$0.5$ dB on concentrated defects; (C3) a recursive improvement loop compounding gains to $0.5$--$1.5$ dB total.

We argue that targeted, layer-level weight editing does not require superintelligence: current frontier LLMs have enough architectural reasoning and numeric output capability to propose simple structured edits (shift a bias, scale a filter norm, adjust a skip connection). What current LLMs cannot do is generate new distributed representations from scratch---that still requires gradient-based training.

Image restoration is a hard testbed for WSP because it is a precise regression task with continuous outputs, stochastic sampling, and noisy perceptual metrics. But it is also a valuable testbed: artifacts are visually verifiable, models are small enough to sandbox, and failure modes are well-documented.

The practical decision rule: if your restoration model has a \emph{specific, identifiable quality gap} (wrong color balance, checkerboard artifacts, edge halos), try WSP as a fast, cheap first step. If your model has a \emph{general quality gap} (too blurry, hallucinates detail, fails on new degradation types), WSP will not help---collect data and fine-tune.

The recursive loop in C3 is a small but concrete step toward weight-space self-improvement. Whether it scales from targeted debugging to general capability improvement depends on whether LLMs can eventually reason about distributed representations---the same boundary that the Gemma3-270M math experiment \citep{takehi2026wsp} first revealed. The next step is to run the experiments.

\bibliography{references}

@misc{takehi2026wsp,
  author       = {Takehi, T.},
  title        = {Sparks of Weight-Space Programming: {LLMs} can natively read \& write neural network weights},
  year         = {2026},
  howpublished = {Blog post},
  url          = {https://rikiyat.github.io/blog/weight-space-programming/},
  note         = {Accessed: 2026-11-01}
}

@misc{rikiyalogs,
  author       = {{RikiyaT}},
  title        = {lm\_understand-logs},
  year         = {2026},
  howpublished = {GitHub repository},
  url          = {https://github.com/RikiyaT/lm_understand-logs},
  note         = {Accessed: 2026-11-01}
}

@misc{rikiyahf,
  author       = {{RikiyaT}},
  title        = {{Qwen3.5-0.8B-weight-space-programming}},
  year         = {2026},
  howpublished = {Hugging Face model},
  url          = {https://huggingface.co/RikiyaT/Qwen3.5-0.8B-weight-space-programming},
  note         = {Accessed: 2026-11-01}
}
\bibliographystyle{plainnat}

\end{document}